\documentclass[conference]{IEEEtran}
\IEEEoverridecommandlockouts
\usepackage{times}
\usepackage{epsfig}
\usepackage{graphicx}
\usepackage{amsmath}
\usepackage{amssymb}
\usepackage{multirow}
\usepackage{booktabs}
\usepackage{url}
\usepackage{xcolor}

\def\BibTeX{{\rm B\kern-.05em{\sc i\kern-.025em b}\kern-.08em
		T\kern-.1667em\lower.7ex\hbox{E}\kern-.125emX}}
\begin{document}

	\title{Multi-Metric Evaluation of Thermal-to-Visual Face Recognition}
	
	\author{\IEEEauthorblockN{Kenneth Lai and Svetlana N. Yanushkevich}
		\IEEEauthorblockA{\textit{Biometric Technologies Laboratory, Dept. Electrical \& Computer Engineering} \\
			\textit{University of Calgary}, Canada\\
			\{kelai, syanshk\}@ucalgary.ca}}

	\maketitle

		\IEEEpubid{\begin{minipage}{\textwidth}\ \\[55pt]
		\footnotesize{{\fontfamily{ptm}\selectfont  Digital Object Identifier 10.1109/EST.2019.8806202 \\978-1-7281-5546-3/19/\$31.00 \copyright 2019 IEEE}}
		\end{minipage}}
	
	\begin{abstract}
		In this paper, we aim to address the problem of heterogeneous or cross-spectral face recognition using machine learning to synthesize visual spectrum face from infrared images. The synthesis of visual-band face images allows for more optimal extraction of facial features to be used for face identification and/or verification. We explore the ability to use Generative Adversarial Networks (GANs) for face image synthesis, and examine the performance of these images using pre-trained Convolutional Neural Networks (CNNs). The features extracted using CNNs are applied in face identification and verification. We explore the performance in terms of acceptance rate when using various similarity measures for face verification.
	\end{abstract} 
	
	\section{Introduction}
	In face biometrics, the ability to extract well-defined facial features is of vital importance. For systems trained using only visual domain-based face images, the feature extraction process is highly specialized for such images. In addition, the performance of face recognition for images taken in low-light or no-light environments is heavily impacted. The use of sensors that operate in other spectra, such as infrared, rather than the visual band has been applied in the past to alleviate this problem.  The infrared spectrum is of particular interest due to its ability to capture ``heat signature'' and is not dependent on lighting. 
	
	Face biometrics is the primary technology in contemporary automated physical access security systems. Current approaches to face verification and identification are reaching outstanding performance, it is heavily restricted to the use of the visual or RGB band of the electromagnetic radiation. Turning focus to other spectra gave rise to infrared face recognition, cross-spectral recognition, and cross-spectrum synthesis. 
	
	Cross-spectral subject recognition leverages existing techniques optimized for visible images. It attempts to create or generate another latent representation in the targeted domain, in order to compare it to the ones obtained from the visible images. An approach that used the polarization-state information was proposed in \cite{gurton2014enhanced,short2015exploiting}. Another approach created multi-spectral images by encoding the magnitude and phase of the Gabor responses \cite{nicolo2012long}.  A method to transform thermal face images into the visible domain using generative adversarial networks (Thermal-to-Visible GAN) is proposed in \cite{zhang2018tv}.
	
	The purpose of cross-spectrum synthesis is to generate images in the desired domain using the information from another domain. The authors of \cite{juefei2015nir} describe a method to reconstruct visible images using near-infrared (NIR) images, and vice versa using joint dictionary learning. A synthesis technique proposed in \cite{riggan2018thermal} uses the global and local regions of the face to create highly discriminative visible-like faces that allow for great verification performance.
	
	An evaluation of using the existing Convolutional Neural Networks (CNNs) trained on visible images to extract features for the purpose of cross-spectral, or heterogeneous, face recognition was reported in \cite{saxena2016heterogeneous}. The experiments with varying metric learning strategies showed that the results from pre-trained CNNs are on par or better than the state-of-the-art techniques developed for NIR images \cite{saxena2016heterogeneous}.
	
	A deep network model proposed by \cite{guo2017face} used both NIR and visual images for face recognition. An adaptive score fusion strategy was used for the final classification, allowing for better performance in combating the problem of illumination in face recognition.

	\section{The Proposed Approach}\label{sec:framework}
	
	In this paper, the goal is to study the feasibility of face verification and identification in the visual spectrum providing only infrared face images as probes. We propose the following approach to reach this goal. We use a Generative Adversarial Network (GAN) to synthesize the visible images from the thermal images, and then apply face recognition techniques on the synthesized images. We start with a set of both the visual and thermal images, and normalize the pairs of such using the facial landmarks. After normalization, we use a GAN to learn and generate the mapping between the thermal and visible images. Finally, we evaluate the performance of face recognition using the synthesized visible images and three different CNNs, namely InceptionV3 \cite{szegedy2016rethinking}, Xception \cite{chollet2017xception}, and MobileNet \cite{howard2017mobilenets}.
	
	\subsection{Database}
	In this paper, the data was taken from the Carl database \cite{espinosa2013new,espinosa2010criterion} which consists of visible and thermal images collected using a thermographic camera TESTO 880-3. The database contains a total of 41 subjects. For each subject, four image acquisition sessions were performed, each with three different lighting settings (natural illumination, infrared illumination, artificial illumination), and five images for each setting. The resolution of the images is 640x480 and 160x120 for the visible and thermal images, respectively.
	
	\subsection{Normalization}
	Since the collection of visible and thermal images both require different sensors, the images taken are generally of different resolution and are misaligned. In this paper, we perform a normalization process that consists of both face alignment and image resizing across the thermal and visual domains.
	
	The normalization process starts with locating the face and then attempts to find the facial landmarks. This stage is only applied to the visual images, as the detector used for facial landmarking is not optimal for thermal images. Due to this challenge, we have developed a mapping function that translates the pixel coordinates from the visual to the thermal image, such that the facial landmarks found in one are imposed onto the other. This approach is explained below.
	
\begin{figure}[!ht]
		\centering
		\includegraphics[width=0.45\textwidth,interpolate]{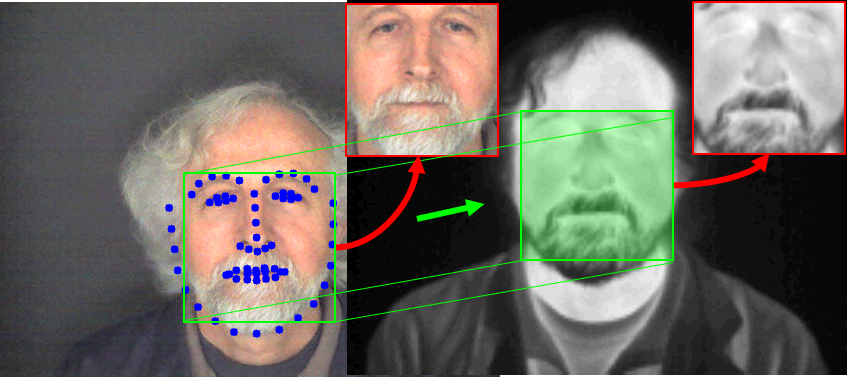}
		\caption{Landmark extraction and centering of a face from the thermal and visible domain.}
		\label{fig:normalization}
	\end{figure}

	We apply a pre-trained detector, \texttt{dlib} \cite{king2009dlib}, that extracts both the 68 landmarks and the location of the face from the visible and thermal images. The extracted faces are centered based on the eyes and nose landmarks, and are resized to the image resolution of 96x96 pixels. Figure \ref{fig:normalization} shows an example of the landmark extraction and centering process that results in the mapping, or alignment of the images in both spectra. 
	
	\subsection{Synthesis}
\begin{figure}[!ht]
		\centering
		\includegraphics[width=0.45\textwidth,interpolate]{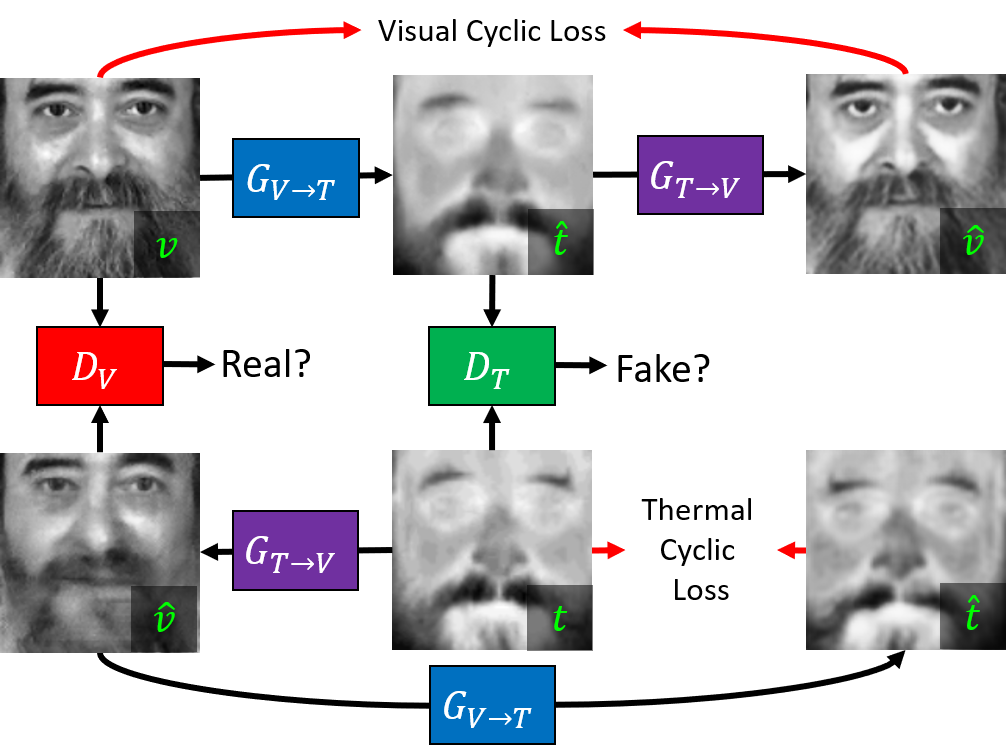} 
		\caption{The architecture of the CycleGAN. $v$ and $t$ are the original visual and thermal images, while $\hat{V}$ and $\hat{T}$ are the synthesized images, respectively.}
		\label{fig:gan}
	\end{figure}

	To generate the synthetic images in the visible spectrum given the infrared images, we propose to use a CycleGAN structure \cite{zhu2017unpaired} to learn the mapping between the images in two different spectra. We modified the CycleGAN structures so it can learn and transform a thermal image into a visible image. One of the major benefits of using CycleGAN for image synthesis is that the pairing (based on the same subject whose images in both spectra are acquired) is not required. It means that the GAN can be trained on a set of thermal images and visual images that are not necessarily taken from the same subjects or in the same setup.  Figure \ref{fig:gan} illustrates the overall CycleGAN structure, consisting of two generators, $G_{V->T}$ and $G_{T->V}$, and two discriminators, $D_{V}$ and $G_{T}$. 
	
\begin{figure*}[!ht]
		\centering
		\includegraphics[width=0.85\textwidth,interpolate]{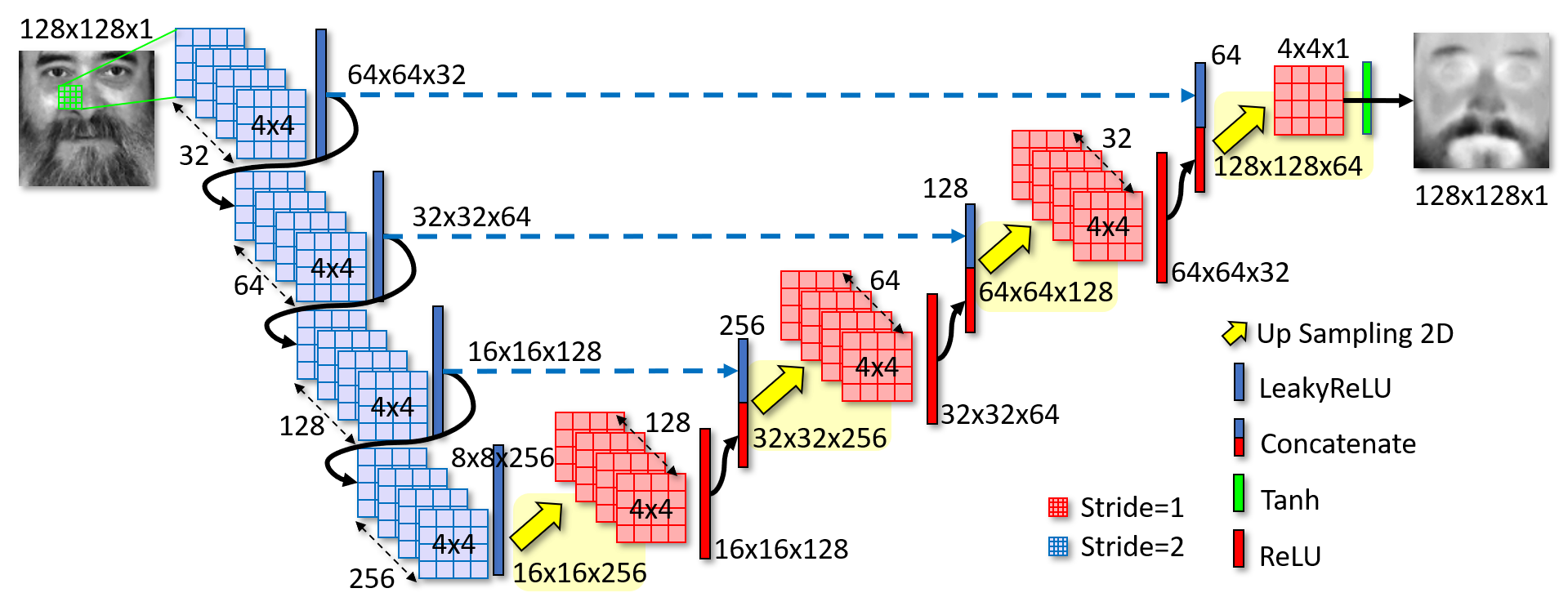} 
		\caption{The generator of the CycleGAN.  The downsampling process is implemented by using  a stride of 2 during the convolution procedure.  Upsampling rescales the input by a factor of 2 using interpolation.  The blue dashed lines represent the skip connections.}
		\label{fig:gen}
	\end{figure*}

	A generic generator, $G_{X->Y}$, enables the process of synthesizing an image in the domain $Y$ from an image in the domain $X$ based on the learned mapping $G: X->Y$. In this paper, the CycleGAN uses two generators, $G_{V->T}$ and $G_{T->V}$, that translates images from the visual domain, $V$, to the thermal domain, $T$, and vice versa. Figure \ref{fig:gen} illustrates the generator architecture of the CycleGAN. The architecture is based on a modified version of the U-Net CNN \cite{ronneberger2015u}. In this paper, we perform convolution with $stride=2$ in order to implement the downsampling processing, as opposed to U-Net which uses $2\times 2$ max pooling. Due to this change, the upsampling process is also slightly modified: the 2D upsampling is performed on the concatenated results between the convolution output and the skip connection.
	
\begin{figure*}[!ht]
		\centering
		\includegraphics[width=0.85\textwidth,interpolate]{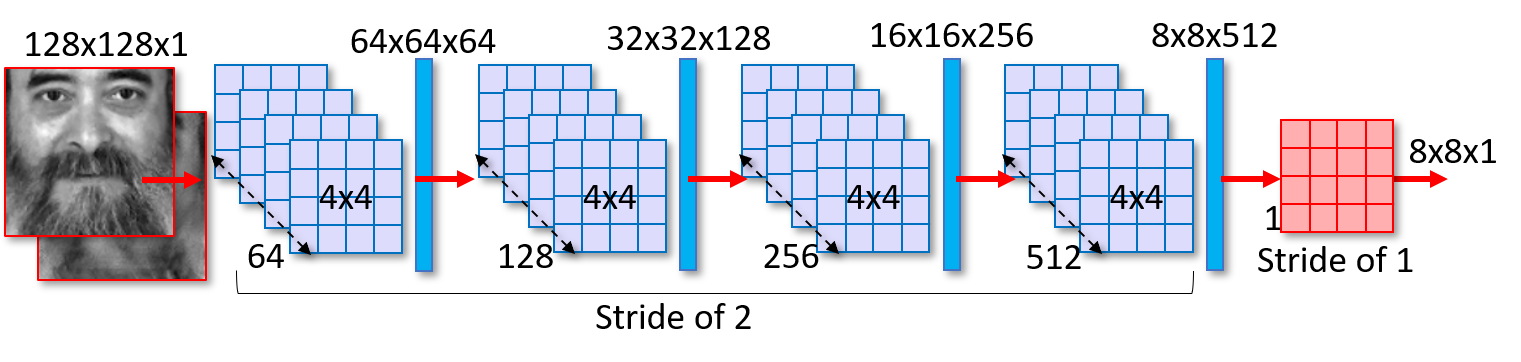} 
		\caption{Discriminator of the CycleGAN.}
		\label{fig:disc}
	\end{figure*}

	A discriminator, $D$, attempts to determine whether the synthesized image $\hat{x}$ is real or fake. This is performed by learning to distinguish between a real image $x$ and a fake image $\hat{x}$. The discriminator chosen in this paper is shown in Figure \ref{fig:disc}. It is based on the patchGAN classifier \cite{li2016precomputed}. The benefit of using a patchGAN classifier is that the output is an $NxN$ matrix, as opposed to a single value (binary Yes or No), as each element in the $NxN$ matrix contains values indicating whether the patch $x_{ij}$ is real or fake.
	
	Together, the generator and the discriminator form two players, $G$ and $D$, of a min-max game where $G$ attempts to generate an image that can fool $D$, while $D$ tries to correctly distinguish between the real and fake images. This process can be interpreted as an adversarial loss \cite{goodfellow2014generative}. In the case of CycleGAN, a new loss called a cycle consistency loss \cite{zhou2016learning}, is introduced that also needs to be minimized. The cyclic loss assumes that given two generators, $G_{V->T}$ and $G_{T->V}$, we are able to execute the approximations $G_{T->V}(G_{V->T}(v))\approx v$ and $G_{V->T}(G_{T->V}(t))\approx t$.
	
	\subsection{Identification}\label{sec:identification}
	
	For identification, we use a four-fold cross-validation process where each fold represents a different acquisition session. Identification is based on a $1:N$ comparison where we attempt to find the identity of the subject. For evaluation, we apply transfer learning on three different CNNs: InceptionV3, Xception, and MobileNet.
	
	We train each network with the set of real (baseline) visual images acquired in three separate sessions, and validate using the images from the remaining session. After selecting the optimal parameters for the particular validation set, we test the final performance of the system with the synthesized visual images. We compare the performance between the baseline and the synthesized images.
	
	The process of transfer learning for each model involves loading the pre-existing weights optimized for the ImageNet challenge. The top fully-connected and classification layers from each model are replaced with an average pooling layer, two fully-connected layers with 512 units, and a classification layer. Two sequential training process is applied to each model. Initially, the model is trained with the bottom layers' weights frozen, allowing only the newly attached classification layers to update at a higher learning rate. After updating the classification layers, the entire model is finely-tuned with a small learning rate.
	
	\subsection{Verification}\label{sec:verification}
	
	For verification, we also apply a four-fold cross-validation process where each fold represents a different acquisition session. Verification is defined as a $1:1$ comparison where we attempt to confirm the claimed identity is correct.  In addition, the evaluation is based on the output of all three networks: InceptionV3, Xception, and MobileNet. In contrast to identification, the verification system uses the networks to generate a feature template for matching.
	
	In this paper, we examine the features obtained from the fully-connected and classification layer. From the fully-connect layer, we can extract 512 features, since there are 512 units in the fully-connected layer, while the classification layer produces 41 score-level features. Using these features as a basis of a template, we compare the similarity between the enrolled and the probe features. We evaluate the performance of common metrics that can be used to compute the similarity between the features. 
	
	Table \ref{tab:metrics} denotes the selected metrics used in this paper for calculating the similarity between the features. The evaluation metrics are based on the distance measures that have been used for comparing different probability distributions. A survey performed in \cite{cha2007comprehensive} shows the common metrics used for calculating similarity/dissimilarity between such distributions. A more exhaustive list of distances can be found in \cite{deza2006dictionary}. In this paper, we have chosen the best performing metric from each family to illustrate their performance.
	
	\begin{table}[!htb]
		\begin{footnotesize}
			\centering
			\caption{Evaluation Metrics}\label{tab:metrics}
			\begin{tabular}{@{}|ll|@{}}
				\hline
				City Block	&	\parbox{6cm}{\begin{equation}	\sum\limits_{i = 1} ^d\left|p_i-q_i\right|	\end{equation}}	\\
				\hline
				Kulczynski d	&	\parbox{6cm}{\begin{equation}	\frac{\sum\limits_{i = 1} ^d\left|p_i-q_i\right|}{\sum\limits_{i = 1} ^d\min(p_i,q_i)}	\end{equation}}	\\
				\hline
				Czekanowski	&	\parbox{6cm}{\begin{equation}	1-\frac{2\sum\limits_{i = 1} ^d\min(p_i,q_i)}{\sum\limits_{i = 1} ^d (p_i+q_i)}	\end{equation}}	\\
				\hline
				Dice	&	\parbox{6cm}{\begin{equation}\label{eq:dice}	1-\frac{2\sum\limits_{i = 1} ^d p_i q_i}{\sum\limits_{i = 1} ^d {p_i}^2 + \sum\limits_{i = 1} ^d {q_i}^2}	\end{equation}}	\\
				\hline
				Squared	&	\parbox{6cm}{\begin{equation}	\sum\limits_{i = 1} ^d \frac{(p_i-q_i)^2}{p_i+q_i}	\end{equation}}	\\
				\hline
				Squared-Chord	&	\parbox{6cm}{\begin{equation}	\sum\limits_{i = 1} ^d (\sqrt{p_i}-\sqrt{q_i})^2	\end{equation}}	\\
				\hline
				Jensen-Shannon	&	\parbox{6cm}{\begin{equation}	\frac{1}{2}\left[\sum\limits_{i = 1} ^d p_i \ln\frac{2p_i}{p_i+q_i} + \sum\limits_{i = 1}^d q_i \ln\frac{2q_i}{p_i+q_i}\right]	\end{equation}}	\\
				\hline
			\end{tabular}
		\end{footnotesize}
	\end{table}

	\section{Experiments and Results}\label{sec:experiments}
	
	To evaluate the performance of CycleGAN used for visual image synthesis for both the identification and verification. A four-fold cross-validation technique is used, and the real visual images are used for training and validation, and the synthesized images are used for testing. 
	
	\subsection{Image Synthesis}
	
\begin{figure}[!ht]
		\centering
		\includegraphics[width=0.48\textwidth,interpolate]{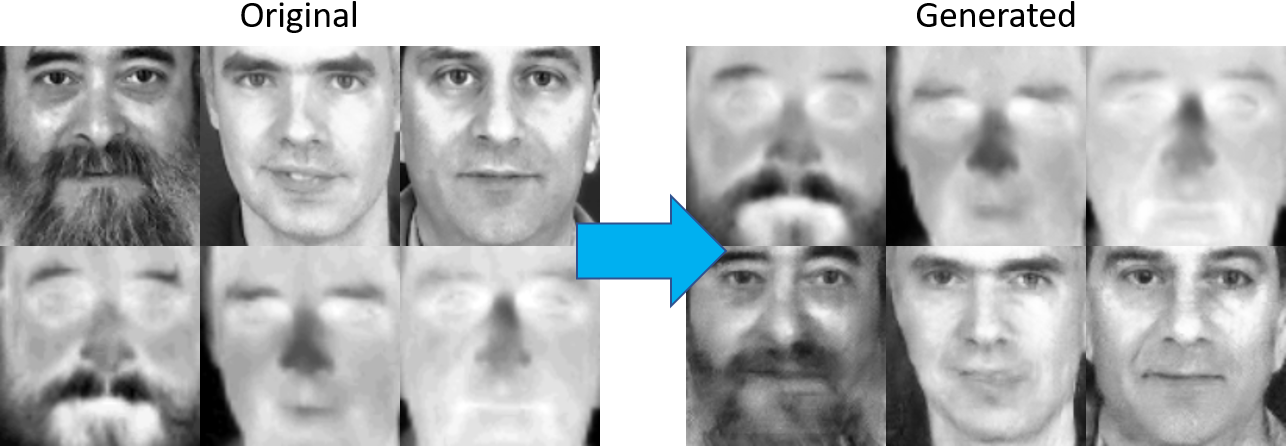} 
		\caption{Samples of real visual and thermal images of three different subjects from the Carl database (left panel), and the generated images using a CycleGAN (right panel).}
		\label{fig:visual}
	\end{figure}

	The original images are taken from the Carl database, cropped and normalized based on the facial landmarks. Each of the thermal and visual images is used to train the CycleGAN such that visual images can be synthesized from the thermal images while the thermal images can be generated from the visual images. Figure \ref{fig:visual} illustrates the original and synthesized faces in both thermal and visual domains.
	
	\subsection{Face Identification}
	
	The performance of face identification ($1:N$) is evaluated in this work using the rank performance and the true acceptance rate at the targeted 1\% and 0.1\% False Acceptance Rate (FAR).  For evaluation, each of the targeted FARs corresponds to a specific acceptance threshold. The testing set was formed based on visual images synthesized from the real thermal images using the GAN trained on both visual and thermal images. The validation set was formed from the original images in the same domain (visual) as the training images.
	
	Table \ref{tab:id} reports the performance for the validation and test sets, for the three network models. Validation set (Valid) shows high and consistent performance because the validation images come from the same domain as the training images. The test set (Test) produces low performance due to the fact that the test sets are the results of synthesis from thermal images. For testing, the Xception model shows the best performance in terms of acceptance rate at the targeted FAR. However, when evaluating the top rank or rank-1 performance, the MobileNet model demonstrates the highest performance. The difference in the performance indicates that Xception provides a wider spread of confident scores which are better for selecting a specific operational point, whereas MobileNet is better for a ranked-based approach.
	
	\begin{table}[!htb]
		\centering
		\begin{footnotesize}
			\caption{Average identification performance (\%) and standard deviation across four-fold cross-validation}\label{tab:id}
			\begin{tabular}{@{}ll||c|c|c@{}}
				\multicolumn{2}{c||}{\multirow{2}{*}{Models}}	&	\multicolumn{2}{c|}{True Acceptance Rate}	&	\multicolumn{1}{c}{\multirow{2}{*}{Rank-1}}	\\
				&& \multicolumn{1}{c|}{@ 1\% FAR} & \multicolumn{1}{c|}{@ 0.1\% FAR} & \\
				\hline
				\hline
				\multirow{3}{*}{\rotatebox[origin=c]{90}{Test}}	&	InceptionV3	&	37.95	$\pm$	1.92	&	14.80	$\pm$	1.54	&	45.03	$\pm$	0.76	\\
				&	Xception	&	\textbf{43.83	$\pm$	1.38}	&	\textbf{20.80	$\pm$	1.34}	&	35.12	$\pm$	2.09	\\
				&	MobileNet	&	28.45	$\pm$	1.37	&	9.90	$\pm$	0.94	&	\textbf{48.91	$\pm$	1.99}	\\
				\hline
				\multirow{3}{*}{\rotatebox[origin=c]{90}{Valid}}	&	InceptionV3	&	\textbf{94.20	$\pm$	2.43}	&	\textbf{85.31	$\pm$	3.18}	&	94.41	$\pm$	1.74	\\
				&	Xception	&	\textbf{94.20	$\pm$	3.04}	&	84.74	$\pm$	6.95	&	91.79	$\pm$	2.87	\\
				&	MobileNet	&	92.53	$\pm$	3.63	&	80.74	$\pm$	6.09	&	\textbf{95.35	$\pm$	1.95}	\\
			\end{tabular}
		\end{footnotesize}
	\end{table}
	
\begin{figure}[!ht]
		\centering
		\includegraphics[width=0.48\textwidth,interpolate]{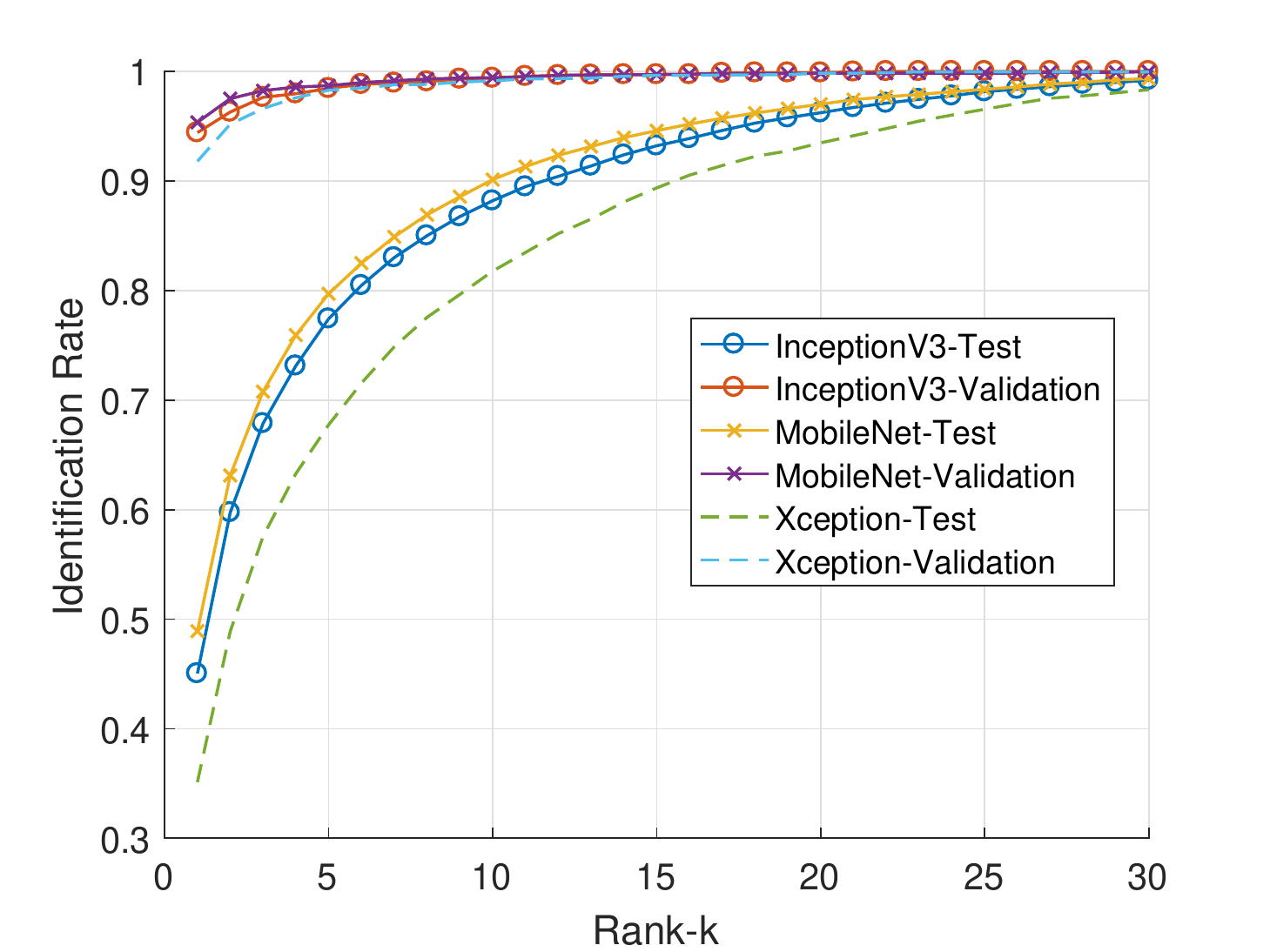}
		\caption{CMC curve for InceptionV3, Xception, and MobileNet using the Carl database. Test case indicates the use of the synthesized visual images, and validation case indicates the use of real visual images.}
		\label{fig:cmc}
	\end{figure}

	Figure \ref{fig:cmc} illustrates the performance of different network models when accepting identities at increasing ranks. The performance is evaluated in terms of the Cumulative Matching Characteristic (CMC) curve which characterizes the identification rate at different rank levels. The CMC curve shows that each of the models operates at a similar performance, with the Xception model performing the worst for the test set.
	
	\subsection{Face Verification}
	For verification, we evaluate the performance using the Receiver Operating Characteristic (ROC) curves, the True Acceptance Rate at targeted FAR, equal error rate (EER), and Area Under the Curve (AUC). First, we examine the performance between the scores obtained from the classification layer and the features extracted from the Fully-Connected (FC) layers. Next, we examine the influence of different similarity metrics for comparing the enrolled and probe feature templates.
	
\begin{figure}[!ht]
		\centering
		\includegraphics[width=0.48\textwidth,interpolate]{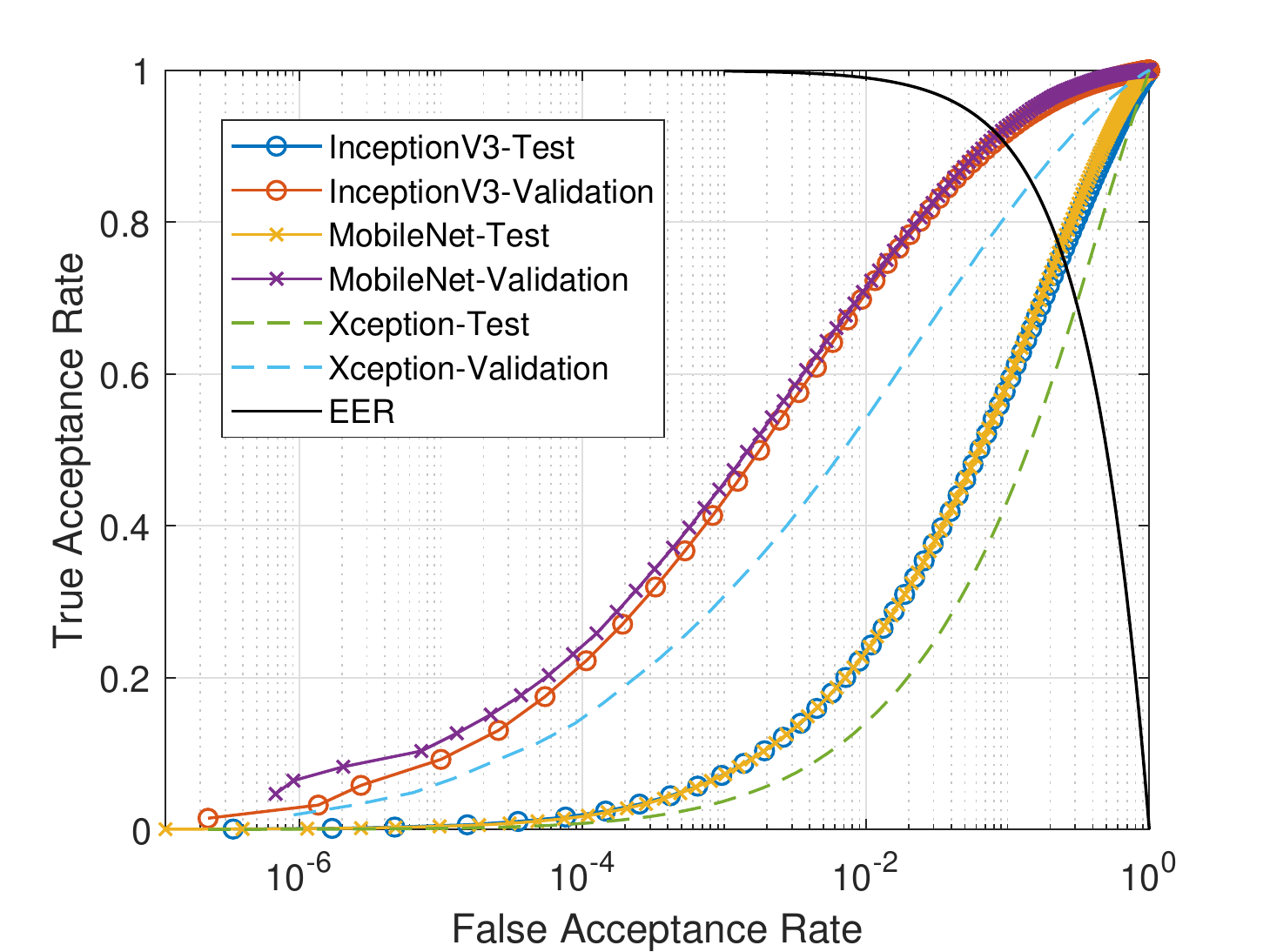}
		\caption{ROC curve using 41 score-level features evaluated using Dice similarity metric.}
		\label{fig:roc41}
	\end{figure}

\begin{figure}[!ht]
		\centering
		\includegraphics[width=0.48\textwidth,interpolate]{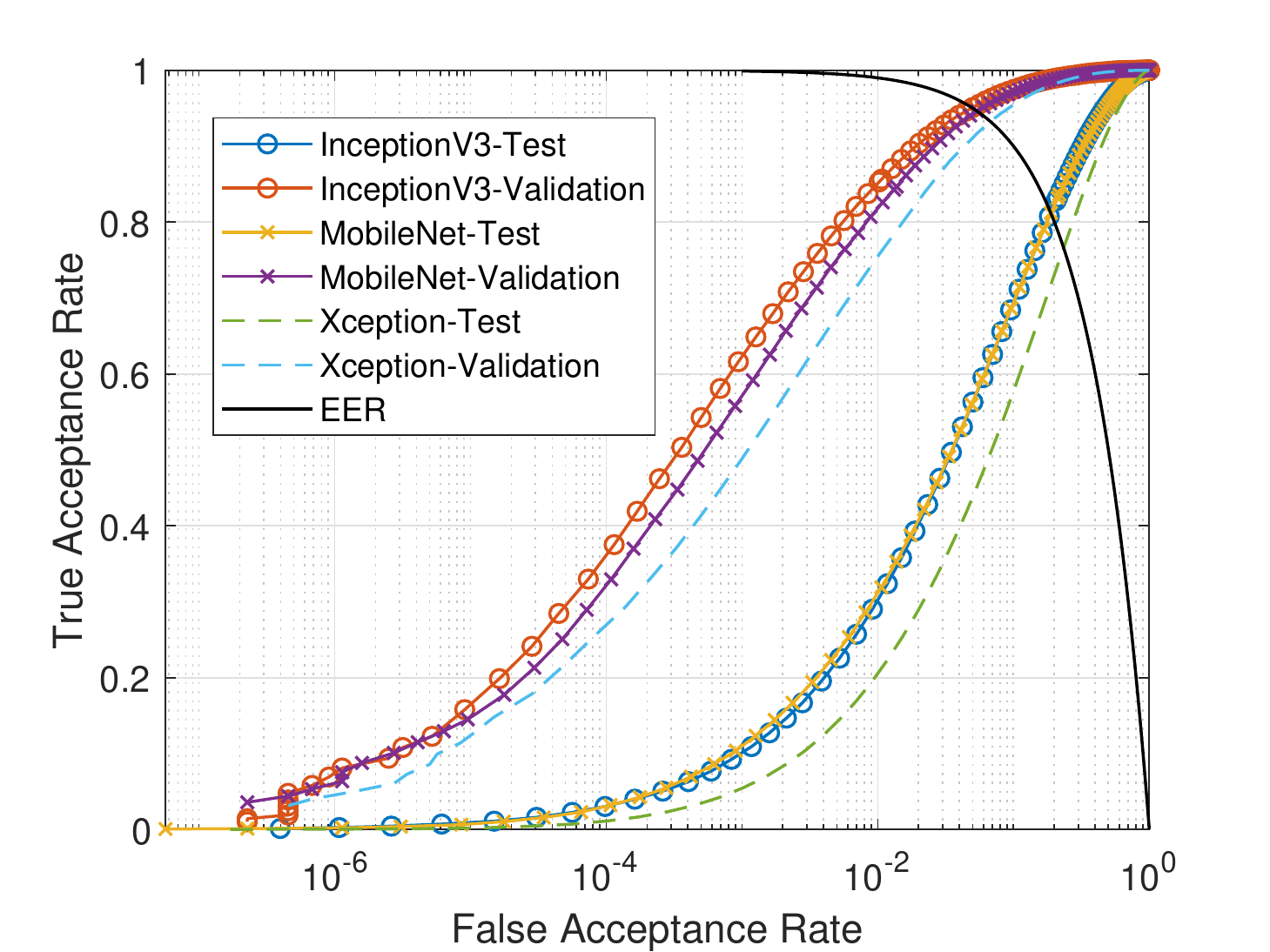}
		\caption{ROC curve created using Dice similarity metric and 512 fully-connected features.}
		\label{fig:roc512}
	\end{figure}

	Verification is a process of matching a probe template to a previously enrolled template ($1:1$ matching). We created two cohorts of ROC curves (Figure \ref{fig:roc41} and \ref{fig:roc512}) for each of the three CNN models: InceptionV3, Xception, and MobileNet, using the Dice metric (Table \ref{tab:metrics}, Equation \ref{eq:dice}). The first group is based on using the score-level features extracted from the classification layer. The second group is generated based on the features from the FC layer. Table \ref{tab:veriCmp} reports the performance between the score-level features and FC features.
	
	As shown in Table \ref{tab:veriCmp}, the FC features outperform the score-level features regardless of the network model, test vs. validation set, and performance metrics. This result indicates that in the process of combining the 512 FC features, the classification layer loses some important information when translating the FC features into the 41 scores. For validation using the score-level features, the Xception model offers the best performance while InceptionV3 slightly outperforms Xception when using FC features. For the testing, both Xception and InceptionV3 offers similar performance for both the score-level and FC features. 
	
	\begin{table*}[!htb]
		\centering
		\begin{footnotesize}
			\caption{Averaged verification performance (\%) and standard deviation using Dice metric across four-fold cross-validation} \label{tab:veriCmp}
			\begin{tabular}{@{}ll||r|r|r|r||r|r|r|r@{}}
				\multicolumn{2}{c||}{\multirow{3}{*}{Models}}		&	\multicolumn{4}{c||}{41 Score-Level Features}		&	\multicolumn{4}{c}{512 Features From the Fully-Connected Layer}											\\
				&& \multicolumn{2}{c|}{True Acceptance Rate} & \multicolumn{1}{c|}{\multirow{2}{*}{EER}} & \multicolumn{1}{c||}{\multirow{2}{*}{AUC}} & \multicolumn{2}{c|}{True Acceptance Rate} & \multicolumn{1}{c|}{\multirow{2}{*}{EER}} & \multicolumn{1}{c}{\multirow{2}{*}{AUC}} \\
				&&\multicolumn{1}{c|}{@ 1\% FAR}&\multicolumn{1}{c|}{@ 0.1\% FAR}&&&\multicolumn{1}{c|}{@ 1\% FAR}&\multicolumn{1}{c|}{@ 0.1\% FAR}&&\\
				\hline																														\hline			
				\multirow{3}{*}{\rotatebox[origin=c]{90}{Test}}	&	InceptionV3	&	23.79	$\pm$	2.24	&	7.46	$\pm$	1.15	&	23.90	$\pm$	1.39	&	83.58	$\pm$	1.68	&	30.25	$\pm$	2.14	&	10.21	$\pm$	1.34	&	18.80	$\pm$	0.52	&	89.37	$\pm$	0.49	\\
				&	Xception	&	23.98	$\pm$	1.18	&	7.36	$\pm$	0.72	&	23.26	$\pm$	1.49	&	84.59	$\pm$	1.69	&	31.16	$\pm$	1.08	&	10.98	$\pm$	1.17	&	18.80	$\pm$	0.65	&	89.25	$\pm$	0.63	\\
				&	MobileNet	&	15.70	$\pm$	0.93	&	4.04	$\pm$	0.28	&	29.32	$\pm$	1.67	&	77.22	$\pm$	2.21	&	20.60	$\pm$	1.30	&	5.44	$\pm$	0.55	&	23.40	$\pm$	1.17	&	84.68	$\pm$	1.18	\\
				\multirow{3}{*}{\rotatebox[origin=c]{90}{Valid}}	&	InceptionV3	&	70.97	$\pm$	5.12	&	44.62	$\pm$	7.64	&	8.73	$\pm$	1.60	&	96.84	$\pm$	1.38	&	85.43	$\pm$	2.55	&	62.90	$\pm$	6.87	&	4.95	$\pm$	0.92	&	98.93	$\pm$	0.50	\\
				&	Xception	&	73.44	$\pm$	5.72	&	47.24	$\pm$	6.82	&	7.90	$\pm$	1.97	&	97.43	$\pm$	1.21	&	82.71	$\pm$	4.74	&	58.85	$\pm$	6.48	&	5.31	$\pm$	1.38	&	98.73	$\pm$	0.63	\\
				&	MobileNet	&	58.45	$\pm$	3.38	&	33.37	$\pm$	2.35	&	13.66	$\pm$	1.99	&	93.10	$\pm$	1.79	&	76.20	$\pm$	4.99	&	49.68	$\pm$	4.00	&	6.74	$\pm$	1.43	&	98.20	$\pm$	0.58	\\
			\end{tabular}
		\end{footnotesize}
	\end{table*}

	Table \ref{tab:idmetrics} reports the best performing metrics from each family using the FC features. As shown in Table \ref{tab:idmetrics}, Dice and Squared-Chord metrics offer on average the best performance for all three network models. Both InceptionV3 and Xception perform similarly, with InceptionV3 performing marginally better on the validation set and Xception performing better on the test set.
	
	\begin{table*}[!htb]
		\centering
		\begin{footnotesize}
			\caption{Averaged verification performance (\%) and standard deviation for various metrics using 512 features from the fully-connected layer across four-fold cross-validation} \label{tab:idmetrics}
			\begin{tabular}{@{}l||r|r|r|r||r|r|r|r@{}}
				\multirow{3}{*}{Metrics}	&	\multicolumn{4}{c||}{Test Set}		&	\multicolumn{4}{c}{Validation Set}											\\
				& \multicolumn{2}{c|}{True Acceptance Rate} & \multicolumn{1}{c|}{\multirow{2}{*}{EER}} & \multicolumn{1}{c||}{\multirow{2}{*}{AUC}} & \multicolumn{2}{c|}{True Acceptance Rate} & \multicolumn{1}{c|}{\multirow{2}{*}{EER}} & \multicolumn{1}{c}{\multirow{2}{*}{AUC}} \\
				&\multicolumn{1}{c|}{@ 1\% FAR}&\multicolumn{1}{c|}{@ 0.1\% FAR}&&&\multicolumn{1}{c|}{@ 1\% FAR}&\multicolumn{1}{c|}{@ 0.1\% FAR}&&\\
				\hline																														\hline		
				\multicolumn{9}{c}{InceptionV3}\\
				City Block	&	27.61	$\pm$	2.79	&	8.80	$\pm$	1.58	&	19.52	$\pm$	0.68	&	88.72	$\pm$	0.64	&	84.42	$\pm$	2.38	&	60.44	$\pm$	6.45	&	4.91	$\pm$	0.86	&	98.91	$\pm$	0.55	\\
				
				Kulczynski d	&	\textbf{30.25	$\pm$	2.16}	&	10.06	$\pm$	1.16	&	18.72	$\pm$	0.51	&	89.46	$\pm$	0.49	&	85.89	$\pm$	2.53	&	63.22	$\pm$	6.44	&	4.88	$\pm$	0.89	&	98.97	$\pm$	0.46	\\
				
				Czekanowski	&	30.22	$\pm$	2.19	&	9.95	$\pm$	1.30	&	18.72	$\pm$	0.51	&	89.47	$\pm$	0.49	&	85.74	$\pm$	2.42	&	63.22	$\pm$	6.44	&	4.89	$\pm$	0.90	&	98.97	$\pm$	0.46	\\
				Dice	&	\textbf{30.25	$\pm$	2.14}	&	\textbf{10.21	$\pm$	1.34}	&	18.80	$\pm$	0.52	&	89.37	$\pm$	0.49	&	85.43	$\pm$	2.55	&	62.90	$\pm$	6.87	&	4.95	$\pm$	0.92	&	98.93	$\pm$	0.50	\\
				
				Squared-Chord	&	29.54	$\pm$	2.66	&	9.35	$\pm$	1.41	&	\textbf{18.30	$\pm$	0.48}	&	\textbf{89.92	$\pm$	0.49}	&	\textbf{86.74	$\pm$	2.64}	&	63.31	$\pm$	6.40	&	4.41	$\pm$	0.90	&	\textbf{99.13	$\pm$	0.42}	\\
				
				Squared	&	29.67	$\pm$	2.72	&	9.67	$\pm$	1.40	&	18.43	$\pm$	0.56	&	89.78	$\pm$	0.52	&	86.48	$\pm$	2.10	&	\textbf{63.90	$\pm$	6.38}	&	4.46	$\pm$	0.90	&	99.10	$\pm$	0.46	\\
				
				Jensen-Shannon	&	29.66	$\pm$	2.70	&	9.48	$\pm$	1.53	&	18.34	$\pm$	0.51	&	89.87	$\pm$	0.50	&	86.66	$\pm$	2.32	&	63.73	$\pm$	6.37	&	\textbf{4.40	$\pm$	0.85}	&	99.12	$\pm$	0.44	\\
				\hline
				\multicolumn{9}{c}{Xception}\\
				City Block	&	24.14	$\pm$	1.05	&	7.17	$\pm$	0.91	&	20.99	$\pm$	0.79	&	87.32	$\pm$	0.84	&	80.55	$\pm$	7.31	&	50.79	$\pm$	9.77	&	5.31	$\pm$	1.56	&	98.77	$\pm$	0.63	\\
				
				Kulczynski d	&	30.86	$\pm$	1.24	&	10.86	$\pm$	1.09	&	18.84	$\pm$	0.74	&	89.24	$\pm$	0.69	&	82.49	$\pm$	4.79	&	58.19	$\pm$	6.40	&	5.37	$\pm$	1.38	&	98.72	$\pm$	0.63	\\
				
				Czekanowski	&	30.85	$\pm$	1.24	&	10.94	$\pm$	1.00	&	18.84	$\pm$	0.74	&	89.24	$\pm$	0.69	&	82.51	$\pm$	4.79	&	58.21	$\pm$	6.40	&	5.35	$\pm$	1.40	&	98.73	$\pm$	0.63	\\
				
				Dice	&	\textbf{31.16	$\pm$	1.08}	&	\textbf{10.98	$\pm$	1.17}	&	18.80	$\pm$	0.65	&	89.25	$\pm$	0.63	&	82.71	$\pm$	4.74	&	\textbf{58.85	$\pm$	6.48}	&	5.31	$\pm$	1.38	&	98.73	$\pm$	0.63	\\
				
				Squared-Chord	&	29.37	$\pm$	1.00	&	9.23	$\pm$	0.83	&	\textbf{18.54	$\pm$	0.65}	&	\textbf{89.65	$\pm$	0.59}	&	84.05	$\pm$	5.28	&	56.45	$\pm$	8.58	&	\textbf{4.60	$\pm$	1.28}	&	\textbf{99.04	$\pm$	0.46}	\\
				
				Squared	&	28.81	$\pm$	1.03	&	9.14	$\pm$	0.84	&	18.97	$\pm$	0.66	&	89.24	$\pm$	0.61	&	84.33	$\pm$	5.83	&	56.98	$\pm$	8.90	&	4.63	$\pm$	1.36	&	99.03	$\pm$	0.50	\\
				
				Jensen-Shannon	&	29.16	$\pm$	0.99	&	9.23	$\pm$	0.83	&	18.73	$\pm$	0.65	&	89.47	$\pm$	0.59	&	\textbf{84.50	$\pm$	5.91}	&	56.80	$\pm$	8.72	&	4.61	$\pm$	1.29	&	\textbf{99.04	$\pm$	0.48}	\\
				\hline
				\multicolumn{9}{c}{MobileNet}\\
				City Block	&	15.27	$\pm$	0.54	&	3.68	$\pm$	0.20	&	25.96	$\pm$	0.73	&	81.93	$\pm$	0.85	&	69.48	$\pm$	3.99	&	39.49	$\pm$	3.72	&	8.05	$\pm$	1.25	&	97.55	$\pm$	0.65	\\
				
				Kulczynski d	&	20.50	$\pm$	1.42	&	\textbf{5.50	$\pm$	0.49}	&	23.35	$\pm$	1.24	&	84.73	$\pm$	1.28	&	76.64	$\pm$	5.10	&	49.89	$\pm$	4.48	&	6.66	$\pm$	1.44	&	98.25	$\pm$	0.58	\\
				
				Czekanowski	&	20.46	$\pm$	1.42	&	5.37	$\pm$	0.60	&	23.35	$\pm$	1.24	&	84.74	$\pm$	1.28	&	76.64	$\pm$	5.08	&	49.89	$\pm$	4.48	&	6.66	$\pm$	1.45	&	98.25	$\pm$	0.58	\\
				
				Dice	&	\textbf{20.60	$\pm$	1.30}	&	5.44	$\pm$	0.55	&	23.40	$\pm$	1.17	&	84.68	$\pm$	1.18	&	76.20	$\pm$	4.99	&	49.68	$\pm$	4.00	&	6.74	$\pm$	1.43	&	98.20	$\pm$	0.58	\\
				
				Squared-Chord	&	19.51	$\pm$	0.96	&	4.87	$\pm$	0.31	&	22.99	$\pm$	1.05	&	\textbf{85.15	$\pm$	1.11}	&	\textbf{78.78	$\pm$	4.27}	&	\textbf{50.25	$\pm$	4.97}	&	\textbf{5.96	$\pm$	1.12}	&	\textbf{98.60	$\pm$	0.43}	\\
				
				Squared	&	19.12	$\pm$	0.83	&	4.84	$\pm$	0.31	&	23.48	$\pm$	0.94	&	84.66	$\pm$	1.03	&	77.92	$\pm$	4.23	&	49.62	$\pm$	4.65	&	6.21	$\pm$	1.19	&	98.46	$\pm$	0.48	\\
				
				Jensen-Shannon	&	19.35	$\pm$	0.93	&	4.86	$\pm$	0.39	&	\textbf{23.19	$\pm$	0.98}	&	84.95	$\pm$	1.07	&	78.44	$\pm$	4.31	&	50.08	$\pm$	4.81	&	6.04	$\pm$	1.15	&	98.54	$\pm$	0.45	\\
				
			\end{tabular}
		\end{footnotesize}
	\end{table*}
	
	\section{Conclusions}\label{sec:conclusions}
	
	This paper proposes to use GANs to generate visual images from thermal images that can be used by a typical CNN for face verification or identification. We used a specific type of GAN, CycleGAN, to generate synthetic visual images from thermal images. We showed we're capable of fine-tuning pre-trained CNNs using the original visual images to recognize synthesized visual images. We examined three common CNNs, Inception V3, Xception, and MobileNet, and achieved a rank-1 face identification accuracy of 95.35\% and 48.91\% on the real and synthesized visual images, respectively. For face verification, an EER of 4.41\% and 18.30\% was achieved, using the Squared-Chord metric on the real and synthesized visual images, respectively.
	
	Further study is required using diverse data sets and better fine-tuning approaches in order to improve the performance of the classifiers.
	
	\section*{Acknowledgment}
	This research was partially supported by the Natural Sciences and Engineering Research Council Canada (NSERC SPG grant ``Biometric-enabled Identity Management and Risk Assessment for Smart Cities'').

\end{document}